\documentclass[10pt,twocolumn,letterpaper]{article}

\usepackage{iccv}
\usepackage{times}
\usepackage{epsfig}
\usepackage{graphicx}
\usepackage{amsmath}
\usepackage{amssymb}
\usepackage{adjustbox}
\usepackage{multirow}
\usepackage{pifont}
\usepackage{booktabs}
\usepackage[numbers,sort,compress]{natbib}

\usepackage[table,xcdraw,dvipsnames]{xcolor}
\usepackage[pagebackref=true, breaklinks=true, colorlinks, citecolor=citecolor, bookmarks=false, pagebackref=true]{hyperref}

\usepackage[capitalize]{cleveref}
\usepackage{fontawesome}

\definecolor{citecolor}{HTML}{0071bc}
\definecolor{frontcolor}{HTML}{325ea5}
\definecolor{backcolor}{HTML}{a58b77}
\definecolor{sidecolor}{HTML}{10768c}
\definecolor{skincolor}{HTML}{dcb7b7}
\definecolor{darkred}{rgb}{0.6, 0.1, 0.05}
\definecolor{DeltaColor}{rgb}{0.039,0.73,0.71}
\definecolor{SigmaColor}{rgb}{0.98,0.45,0.0}
\definecolor{AlphaColor}{rgb}{0,0,0.8}
\definecolor{BetaColor}{rgb}{0.8,0,0.8}
\definecolor{GammaColor}{rgb}{0.514,0.34,0.224}
\definecolor{EpsilonColor}{rgb}{0.353,0.725,0.906}
\definecolor{PurpleColor}{HTML}{8B008B}
\definecolor{BadColor}{HTML}{C0392B}
\definecolor{OrangeColor}{rgb}{0.914,0.541,0.0.141}
\definecolor{GreenColor}{rgb}{0.137,0.573,0.565}
\definecolor{RedColor}{rgb}{0.949,0.275, 0.224}
\definecolor{LightCyan}{rgb}{0.88,1,1}
\definecolor{Gray}{gray}{0.85}
\definecolor{black}{gray}{0}

\crefname{section}{Sec.}{Secs.}
\Crefname{section}{Section}{Sections}
\Crefname{table}{Table}{Tables}
\crefname{table}{Tab.}{Tabs.}

\newlength\savewidth\newcommand\shline{\noalign{\global\savewidth\arrayrulewidth
  \global\arrayrulewidth 1pt}\hline\noalign{\global\arrayrulewidth\savewidth}}

\renewcommand{\etal}{\mbox{et al.}\xspace}

\newcommand{\yihao}[1]{{\color{black} #1 }}

\newcommand{\qheading}[1]{\noindent\textbf{#1.}}

\DeclareTextFontCommand{\specific}{\small\fontfamily{qcr}\selectfont}

\newcommand{\page}{\href{https://github.com/psyai-net/D-IF_release}{\specific{github.com/psyai-net/D-IF\_release}}\xspace}

\definecolor{bestcolor}{rgb}{1, 0.5, 0.25}
\definecolor{secondbestcolor}{rgb}{1, 0.8, 0.5}
\newcommand{\bone}{\cellcolor{bestcolor}}
\newcommand{\btwo}{\cellcolor{secondbestcolor}}

\iccvfinalcopy 

\def\iccvPaperID{2638} 

\begin{document}

\title{D-IF: Uncertainty-aware Human Digitization via Implicit Distribution Field}

\author{
Xueting Yang$^{1\dag}$ \quad Yihao Luo$^{1,2\dag}$ \quad Yuliang Xiu$^3$ \quad Wei Wang$^1$ \quad Hao Xu$^{1,4}$ \quad Zhaoxin Fan$^{1,4*}$\\
{\normalsize $^1$Psyche AI Inc. \quad $^2$Imperial College London \quad $^3$Max Planck Institute for Intelligent Systems}\\
{\normalsize $^4$Hong Kong University of Science and Technology}\\
{\tt\small yangxueting@psyai.net \quad y.luo23@imperial.ac.uk \quad yuliang.xiu@tue.mpg.de} \\ 
{\tt\small faithwwei@bupt.edu.cn \quad \{hxubl,zfanaq\}@connect.ust.hk}\\
}

\maketitle


\ificcvfinal\thispagestyle{empty}\fi

\begin{abstract}
   
   Realistic virtual humans play a crucial role in numerous industries, such as metaverse, intelligent healthcare, and self-driving simulation. But creating them on a large scale with high levels of realism remains a challenge. The utilization of deep implicit function sparks a new era of image-based 3D clothed human reconstruction, enabling pixel-aligned shape recovery with fine details. Subsequently, the vast majority of works locate the surface by regressing the deterministic implicit value for each point. However, should all points be treated equally regardless of their proximity to the surface? In this paper, we propose replacing the \textbf{implicit value} with an \textbf{adaptive uncertainty distribution}, to differentiate between points based on their distance to the surface. This simple ``value $\Rightarrow$ distribution'' transition yields significant improvements on nearly all the baselines. Furthermore, qualitative results demonstrate that the models trained using our uncertainty distribution loss, can capture more intricate wrinkles, and realistic limbs. Code and models are available for research purposes at \page.
\end{abstract}

\def\thefootnote{\dag}\footnotetext{These authors contributed equally to this work.}
\def\thefootnote{*}\footnotetext{represents the corresponding author.}

\section{Introduction}


The creation of realistic digital avatars with intricate clothing details holds significant importance in various applications, such as metaverse~\cite{xiang2022dressingavatar}, intelligent healthcare~\cite{RCareWorld}, teleportation~\cite{li2020monoportRTL}, and self-driving simulation~\cite{yang2021s3}. However, conventional methods require substantial human resources and considerable costs for designing or capturing high-fidelity 3D digital avatars. To simplify this process while maintaining quality, both the academic community and industry have shifted their focus and efforts toward data-driven approaches~\cite{saitoPIFuPixelAlignedImplicit2019,li2020monoport,xiuICONImplicitClothed2022}, to accurately reconstruct 3D humans from images or monocular videos.

\begin{figure}[t]
\begin{center}
   \includegraphics[width=1.0 \linewidth]{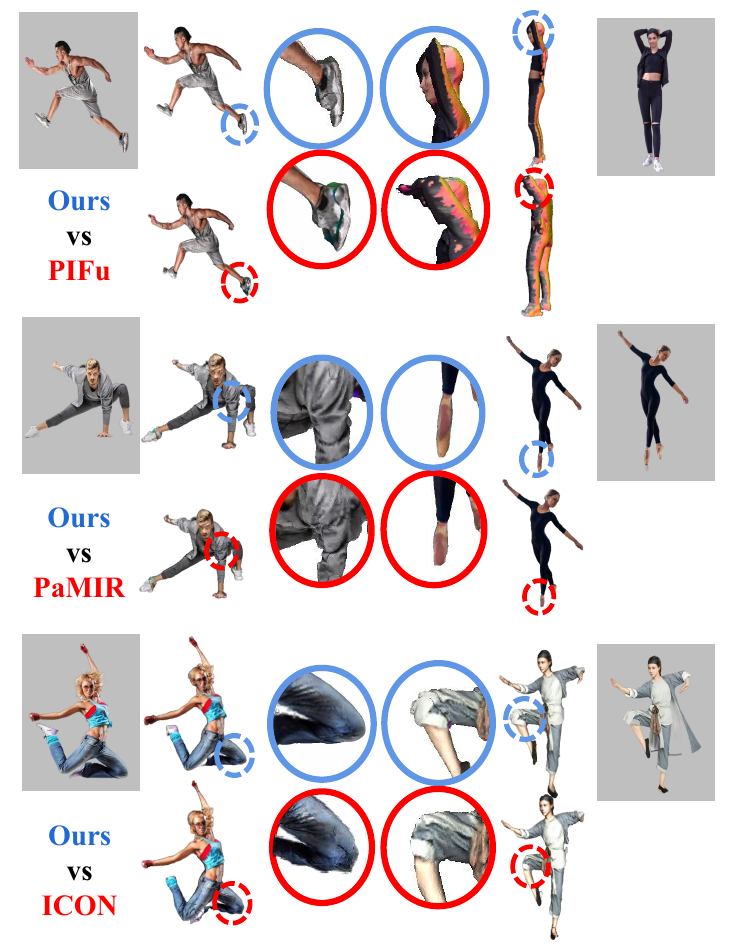}
\end{center}
\vspace{-0.5 em}
\caption{\textbf{Comparison with mainstream SOTAs}. Unlike PaMIR~\cite{zhengPaMIRParametricModelConditioned2020}, PIFu~\cite{saitoPIFuPixelAlignedImplicit2019}, and ICON~\cite{xiuICONImplicitClothed2022}, which frequently produce 3D humans with distorted or non-human shape limbs, missing details, and high-frequency noise, our method overcomes these issues and achieves superior geometric details in reconstruction.}
\label{fig:0}
\vspace{-1.0 em}
\end{figure}

\begin{figure*}[t]
\begin{center}
   \includegraphics[width=1.0\linewidth]{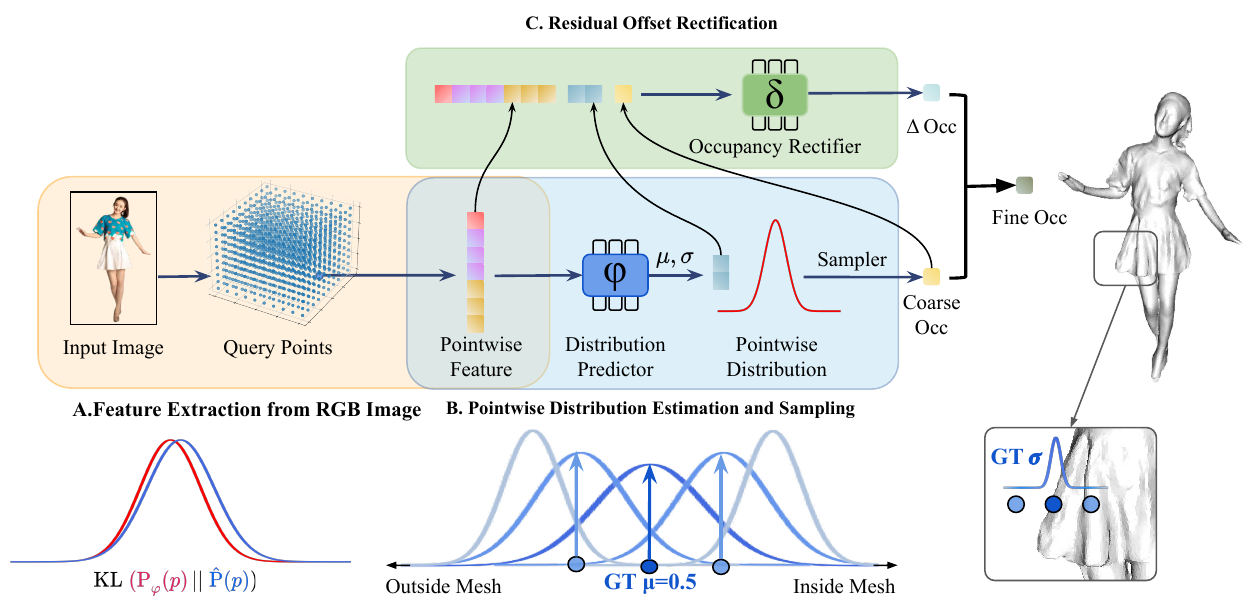}
\end{center}
\caption{\textbf{The framework of D-IF (\cref{sec: distribution-guided implicit function learning network}).} \textbf{A.} For queried points $p \in \mathbb{R}^3$, which are uniformly sampled from the entire 3D space, 7D pointwise local feature $F_{\rm 7D}$ is extracted. 
\textbf{B.} $F_{\rm 7D}$ is then fed into distribution predictor $\mathit{P}_{\varphi}\left(p\right)$, to estimate the per-point distribution ($\mu,\sigma$), and the coarse occupancy $\mathit{O}_s(p)$ is sampled from it. 
\textbf{C.} Given all above features and outputs (9D), Occupancy Rectifier, an additional MLP, is to predict the residual offset  $\Delta \mathit{O}_s(p)$. Finally, the fine occupancy $\widetilde{\mathit{O}_s}(p)$ is obtained via $\widetilde{\mathit{O}_s}(p) = \mathit{O}_s(p) + \Delta \mathit{O}_s(p)$. The bottom half of this illustration demonstrates the design of uncertainty-aware supervised learning (\cref{sec: uncertainty distribution loss}). During training, we formulate the ``pseudo'' ground-truth distribution $\hat{P}\left(p\right)$ as follows: ground-truth smooth occupancy value $\mathit{O}_{gt}(p)$ as the expectation $\mu$ (\cref{def：Smooth Occupancy}), and $\sigma$ is gradually reducing as the point-to-mesh distance increases (\cref{eq:gt-sigma}). Finally, KL-divergence loss is introduced to minimize the difference between predicted and pseudo distributions, see \cref{eq:kl-loss}.}
\label{fig:pipeline}
\end{figure*}

Explicit-shape-based approaches~\cite{alldieck2019tex2shape, alldieck2019learning, xiang2020monoclothcap, lazova2019360,alldieck2018detailed,zhu2019detailed} typically fit and deform the parametric body model, like SMPL-(X)~\cite{loper2015smpl,pavlakos2019expressive}, to align with the visual observations. While these approaches could recover 3D humans wearing tight clothing, they face limitations when it comes to reconstructing loose-fitting garments that largely deviate from the body. 
Alternatively, implicit-function-based approaches~\cite{heGeoPIFuGeometryPixel, saitoPIFuHDMultiLevelPixelAligned2020, zhengPaMIRParametricModelConditioned2020, xiuICONImplicitClothed2022, xiuECONExplicitClothed2022, alldieckPhotorealisticMonocular3D2022,liao2023car,heARCHAnimationReadyClothed2021, hongStereoPIFuDepthAware2021} utilize implicit function parameterized by MLPs to regress occupancy fields~\cite{mescheder2019occupancy} or signed distance fields (SDF)~\cite{park2019deepsdf}. Detailed meshes can then be extracted using Marching Cubes~\cite{lorensen1987marching} from the iso-surface of a certain implicit value. Implicit methods have demonstrated their superiority in capturing geometric details and accommodating various topological structures. However, they may generate non-human shapes for unseen poses or garments, due to the absence of shape regularization.
%


Despite the impressive results of prior implicit-based methods, they have not fully taken into account the presence of \textit{``uncertainty''} in the geometric deformation that arises during the reconstruction procedure. For example, in the case of points that significantly deviate from the body, a practical shortcut would be to categorize them as ``outside points'', while even a minor estimation disturbance near the surface can lead to a completely wrong occupancy result.

To account for such uncertainty, this paper introduces implicit distribution fields, called D-IF. Instead of directly estimating the implicit value at each point, we opt to sample the implicit value from an estimated distribution. This enables us to effectively capture the uncertainty associated with the distance from the surface.
The overview of our method is shown in~\cref{fig:pipeline}. Inspired by ICON~\cite{xiuICONImplicitClothed2022}, we firstly extract 7D local features from the input image and estimated SMPL body. These features are then utilized to estimate the point-wise occupancy distribution. Sampling from the projected distribution of grid points across the entire 3D space produces a coarse occupancy field. To enhance the level of detail, we introduce an additional MLP called ``Occupancy Rectifier'', which refines the coarse occupancy field further, to obtain the fine occupancy field, the final clothed mesh is extracted using Marching Cubes~\cite{lorensen1987marching} at 0.5 level-set.


Upon delving deeper into D-IF, it becomes apparent that there exists a dilemma for the learned distribution to be simultaneously accurate and uncertain. This necessitates finding a balance in terms of distribution sharpness. To address this dilemma, we introduce an explicit supervision mechanism known as the uncertainty distribution loss to learn the distribution, which is illustrated in~\cref{fig:pipeline}.
%
The insight behind the loss is based on the assumption that point-wise distribution is highly relevant to point-to-mesh distance. We elaborate this in ~\cref{sec: uncertainty distribution loss}, where we introduce the KL-divergence~\cite{liese2006divergences} between the predicted distribution and a designed distribution (pseudo GT). Moreover, the Occupancy Rectifier module aims to correct any erroneous occupancy while simultaneously refining intricate shape details.

Quantitative experiments on CAPE~\cite{ma2020learning} confirm that our method achieves SOTA performance, see~\cref{table:1}. And as a ``plug-and-play'' module, it significantly improves the reconstruction accuracy on nearly all the mainstream baselines. As depicted in~\cref{fig:0}, D-IF excels at recovering intricate geometric features, while mitigating common artifacts found in other works~\cite{saitoPIFuPixelAlignedImplicit2019, zhengPaMIRParametricModelConditioned2020, xiuICONImplicitClothed2022}, including non-human parts, missing details, and high-frequency noise.

\section{Related Work}

\subsection{Explicit-shape based human reconstruction}



Parametric models have been widely used in 3D human reconstruction. Previous works~\cite{alldieck2019learning, lazova2019360, alldieck2019tex2shape} have introduced the concept of SMPL+D, where displacement is added to the vertices of the SMPL~\cite{loper2015smpl} model to represent clothed models. For example, Tex2Shape~\cite{alldieck2019tex2shape} defines the vertex offset in the UV space of SMPL to achieve higher-resolution representations with detailed clothing wrinkles. While MGN~\cite{bhatnagar2019multi} performs vertex segmentation on SMPL for different clothing types, enabling better expression of clothing boundaries in reconstructed SMPL+D representations. Alternative parametric methods, inspired by the representation of SMPL+D, propose vertex deformations on parametric SMPL models to capture more geometric details. For instance, Zhu \etal~\cite{zhu2019detailed} employ hierarchical free-form 3D deformation to improve the geometry of the predicted human body. While Weng \etal~\cite{weng2019photo} propose using normal directions to improve deformations and obtain better clothed human body meshes from the SMPL model. Though these methods achieve acceptable performance, there are limitations regarding their ability to express various clothing types due to inherent topology constraints imposed by parametric models. Additionally, learning geometric details from explicit parametric models can be challenging.

\subsection{Implicit-based human reconstruction}

Implicit representations are employed in clothed human reconstruction~\cite{maSCALEModelingClothed2021, huangARCHAnimatableReconstruction2020, heGeoPIFuGeometryPixel,alldieckPhotorealisticMonocular3D2022, xiuICONImplicitClothed2022, chenSNARFDifferentiableForward2021, heARCHAnimationReadyClothed2021, xiuECONExplicitClothed2022} to overcome the constraints associated with parametric representations, which have emerged as the prevailing methods. Among these methods, PIFu~\cite{saitoPIFuPixelAlignedImplicit2019} is the first to employ pixel-aligned features to regress the occupancy field. PIFuHD~\cite{saitoPIFuHDMultiLevelPixelAligned2020} utilizes a coarse and fine network structure and additional normals as additional geometric information to improve PIFu. PaMIR~\cite{zhengPaMIRParametricModelConditioned2020} employs SMPL shape prior as shape regularization. ICON~\cite{xiuICONImplicitClothed2022} and ECON~\cite{xiuECONExplicitClothed2022} still utilize the SMPL-(X) body prior but focus on improving the pose robustness and garment topological flexibility respectively. In contrast to the above-mentioned methods, which are limited to static mesh outputs which are not ready for animation, other methods such as ARCH~\cite{huangARCHAnimatableReconstruction2020}, ARCH++~\cite{heARCHAnimationReadyClothed2021}, aim to reconstruct 3D humans in canonical space.

While the methods mentioned above have demonstrated satisfactory outcomes in cloth reconstruction, they depend on deterministic predictions of implicit values for each point and disregard the significance of incorporating uncertainties inherent in the cloth reconstruction process.

\subsection{Distribution-based 3D reconstruction}


In recent years, research on distribution-based implicit functions has drawn attention. For instance, MaGNet~\cite{baeMaGNetMultiViewDepth2022} estimates a probability distribution of single-view depth, achieving higher accuracy yet evaluating fewer depth candidates. This approach is similar to the one proposed in~\cite{roessle2022dense}. SubFocal~\cite{chao2022learning} estimates the Dirac delta distribution within the range of pixel depth for the input image and uses it as a backpropagation supervisor to reduce bad-pixels. Additionally, CaDDN~\cite{reading2021categorical} projects rich contextual features into the appropriate depth interval in 3D space using a projected categorical depth distribution for each pixel. Distribution-based methods have also been applied to point cloud completion and pose estimation tasks, where they learn distributions for shape completion or motion changes~\cite{rempeHuMoR3DHuman2021, panVariationalRelationalPoint2021}.

Although the above methods have achieved significant improvements, most existing approaches tend to overlook the variations in distribution among different spatial points. In this study, we propose not only a distribution to express the uncertainty of clothing but also a method to differentiate between near-surface and floating points through the utilization of the proposed uncertainty distribution loss. To our knowledge, this study represents the first attempt to address the uncertainties related to clothed human reconstruction.


\section{Method}

Our study aims to generate detailed human meshes, including clothes and hair, from performer images using implicit distribution fields. \cref{fig:pipeline} illustrates an overview of our method. In our work, the clothed human mesh is represented by smooth occupancy (\cref{sec: representing mesh with smooth occupancy}). First, we elaborate on our observations about uncertainty (\cref{sec: An observation of Uncertainty Corresponding to Location }). In particular, a simple yet effective way is proposed to use a distribution-guided implicit network to learn the implicit distribution of each query point (\cref{sec: distribution-guided implicit function learning network}), named D-IF. To train D-IF, we further propose an uncertainty distribution loss to constrain the predicted distribution (\cref{sec: uncertainty distribution loss}). Detailed information on the key designs of our approach is presented in the following sections.

\subsection{Learning the implicit distribution filed}
\label{sec: distribution-guided implicit function learning network}

\yihao{
The aim of this work is to reconstruct a 3D human mesh with details on clothes from a single image. Further than predicting the point-wise value of implicit fields like the occupancy field~\cite{mescheder2019occupancy} and the signed distance field (SDF)~\cite{park2019deepsdf}, we tend to infer the probability distribution of the implicit value for every point by a neural network, which will be proven to maintain appropriate uncertainty but improve the reconstruction accuracy. The point-wise sampling of the predicted implicit distribution fields will regress a classical implicit field to represent the target surface.

We extract the same local deep features as ICON~\cite{xiuICONImplicitClothed2022}, and use them to predict the implicit distribution of each queried point. The extracted features include SMPL-body surface normal $N_{\rm body}\in\mathbb{R}^3$, clothed surface normal $N_{\rm clothed}\in\mathbb{R}^3$, and the signed distance value to the SMPL body $\text{SDF}_{\rm body}\in\mathbb{R}$. The final 7D pixel-aligned feature $\mathit{F}_{{\rm 7D}}(p)$ is constructed by concatenating ($\oplus$) $N_{\rm body}, N_{\rm clothed}$ and $\text{SDF}_{\rm body}$:
\begin{equation}
    \mathit{F}_{{\rm 7D}}(p) := N_{\rm body}\oplus N_{\rm clothed} \oplus \text{SDF}_{\rm body},
\end{equation}

While taking $\mathit{F}_{{\rm 7D}}(p)$ as inputs to predict directly the deterministic occupancy value is a viable option, as previously discussed, uncertainty should also be considered. To this end, we present a framework D-IF involving learnable implicit distribution fields to predict point-wise occupancy values with uncertainty. Specifically, for each query point $p$, an MLP designed as the Distribution Predictor in D-IF is trained to learn the distribution ${P}_{\varphi}(\mathit{F}_{{\rm 7D}}(p))$ which is assumed to be a Gaussian distribution~\cite{bishop2006pattern}: 
\begin{align}
\begin{split}
&\mathit{O}_{s}(p) \sim {P}_{\varphi}(\mathit{F}_{{\rm 7D}}(p))=\mathcal{N}\left(\mu_{\varphi}\left(p\right), \sigma_{\varphi}\left(p\right)\right)\\
 &f(\mathit{O}(p)=Y) =\frac{1}{\sigma_{\varphi}\left(p\right) \sqrt{2 \pi}} e^{-\frac{1}{2}\left(\frac{Y-\mu_{\varphi}\left(p\right)}{\sigma_{\varphi}\left(p\right)}\right)^2}.
\end{split}
\end{align}
where probability density function $f$ of the occupancy value formulates the conventional Gaussian distribution $\mu_{\varphi}\left(p\right)$ and $\sigma_{\varphi}\left(p\right)$ are the mean and the variance of the predicted distribution at the query point $p$. Through the learning of point-wise distribution, the coarse-level occupancy $\mathit{O}_{s}(p)$ will be sampled from the learned distribution.


Coarse-level occupancy from sampling maintains the uncertainty but lacks the accuracy to reconstruct the ground-truth mesh. To keep the balance between uncertainty and accuracy, D-IF designs an additional MLP as the Occupancy Rectifier $R_\delta$ to modify the coarsely sampled value into a fine result. The Occupancy Rectifier will concatenate ($\oplus$) the 7D features $F_{\rm 7D}$, the mean $\mu_{\varphi}\left(p\right)$ and variances $\sigma_{\varphi}\left(p\right)$ predicted from $P_\varphi(F_{\rm 7D}(p))$, and the sampled coarse occupancy $\mathit{O}_{s}(p)$ as input. All these features will be used to estimate the residual $\Delta \mathit{O}_{s}(p)$, which will be added onto the coarse occupancy $\mathit{O}_{s}(p)$, and get the fine occupancy results $\widetilde{\mathit{O}_{s}}(p)$ by the small amendments,
\begin{equation}
\begin{aligned}
\Delta \mathit{O}_{s}(p) &= R_\delta \left( \mathit{O}_{s}(p) \oplus P_\varphi(F_{\rm 7D}(p))  \oplus F_{\rm 7D}(p)\right),\\
\widetilde{\mathit{O}_{s}}(p) &= \mathit{O}_{s}(p) + \Delta \mathit{O}_{s}(p), 
\end{aligned}
\end{equation}
 where Occupancy Rectifier is a MLP parameterized by $\delta$.
}



\begin{figure}[h]
\begin{center}
   \includegraphics[width=1.0\linewidth]{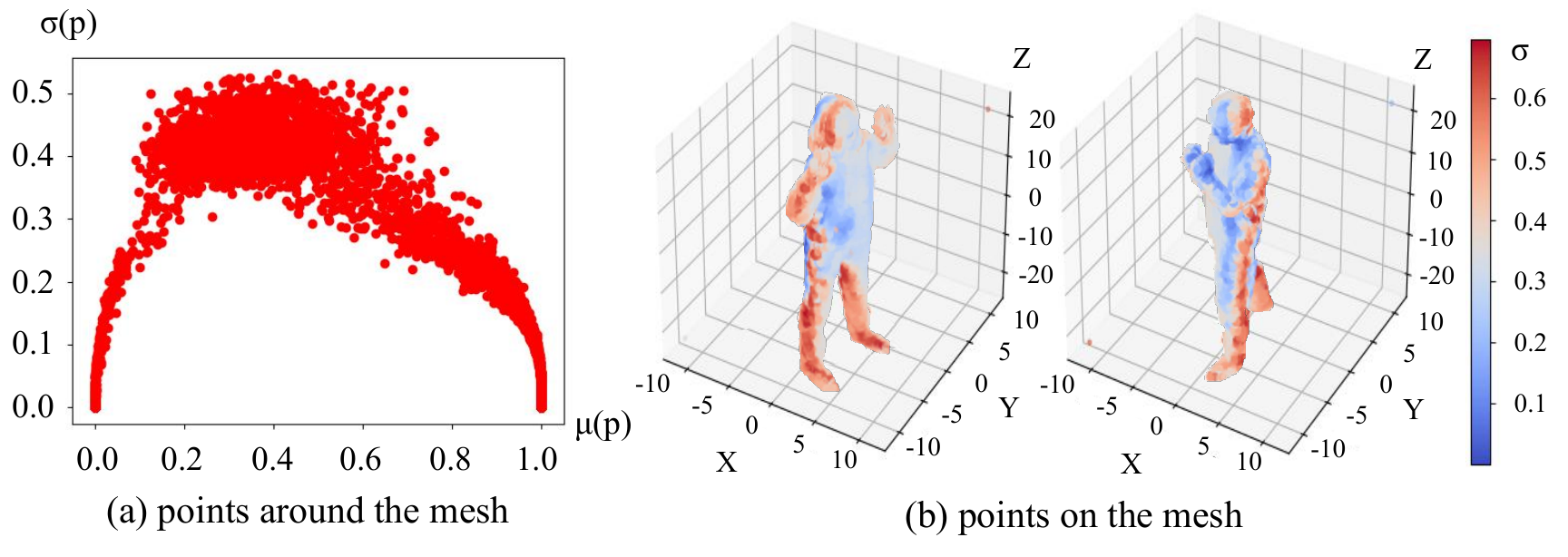}
\end{center}
   \caption{\textbf{Spatial-aware Uncertainty.} For a) points around the mesh and b) points on the surface. The degree of uncertainty increases as points approach the surface, and the body sides, which are typically with higher articulation than the torso.}
\label{fig:uncertaintypic}
\end{figure}

\subsection{Discussion on spatial-aware uncertainty}
\label{sec: An observation of Uncertainty Corresponding to Location }
\yihao{
%
In line with traditional Bayesian deep learning principles \cite{neal1995bayesian,denker1990transforming}, we adopt a two-fold perspective on uncertainty during the prediction of the implicit field. Firstly, there exists \textit{epistemic (model) uncertainty} \cite{graves2011practical} stemming from the idealized network architecture design and incomplete supervision through blend loss functions. This form of uncertainty signifies an inherent limitation which is not specifically addressed in our method. However, our primary focus lies on \textit{aleatoric (data) uncertainty} derived from the data itself ~\cite{kendall2017uncertainties}, reflecting random turbulence in geometric details present on a clothed human shape. Rather than homoscedastic uncertainty with constant strength irrespective of data properties, we emphasize heteroscedastic uncertainty that varies based on input data. Moreover, we contend that the uncertainty associated with the implicit value --- specifically, the variance within the predicted distribution --- is influenced by the spatial relationship between a query point and the potentially reconstructed surface.

Our experiments support this assumption. Specifically, we incorporate the data-dependent Bayesian Loss~\cite{kendall2017uncertainties} to facilitate the learning of heteroscedastic uncertainty:
\begin{equation}
    \mathcal{L}_{B} = \frac{1}{2N}\sum_{i=1}^N\frac{1}{\sigma(p_i)^2} {\|\hat{\mathit{O}}(p_i) -\mathit{O}(p_i)\|}^2 + \log{\sigma(p_i)},
\end{equation} 
where the first term measures the Mahalanobis distance~\cite{mahalanobis2018generalized} between the ground truth value and the value sampled from the predicted distribution, while the last regularization term of variance $\sigma$ prevents the uncertainty from exploding. When the D-IF was only trained with the data-dependent Bayesian Loss, the variance at query points displays a declining tendency as the distance from the point to the target surface increases. Please find the illustration and explanation of the ``spatial-aware uncertainty'' from~\cref{fig:uncertaintypic}

Based on the above assumption and observations, we propose a loss design to model the uncertainty, termed the \textbf{uncertainty distribution loss}, which enables effective supervision of the variance in implicit distribution fields.
}

\begin{table*}[ht]
\centering
\resizebox{\linewidth}{!}{
\begin{tabular}{lc|c|c|ccc|ccc|ccc}
 &
   &
   &
   &
  \multicolumn{3}{c|}{CAPE-FP} &
  \multicolumn{3}{c|}{CAPE-NFP} &
  \multicolumn{3}{c}{CAPE} \\ 
 &
  Methods &
  \begin{tabular}[c]{@{}c@{}}Smooth \\ Occupancy\end{tabular} &
  \begin{tabular}[c]{@{}c@{}}Uncertainty \\ Dist. Loss\end{tabular} &
  Chamfer $\downarrow$ &
  P2S $\downarrow$ &
  Normals $\downarrow$ &
  Chamfer $\downarrow$ &
  P2S $\downarrow$ &
  Normals $\downarrow$ &
  Chamfer $\downarrow$ &
  P2S $\downarrow$ &
  Normals $\downarrow$ \\ \shline
\multicolumn{1}{c}{\multirow{4}{*}{A}} &
  Ours &
  $\checkmark$ &
  $\checkmark$ &
  \bone 0.684 &
  \bone 0.677 &
  \btwo 0.048 &
  \bone 0.838 &
  \bone 0.821 &
  \btwo 0.055 &
  \bone 0.785 &
  \bone 0.771 &
  \btwo 0.050 \\ 
  \cline{1-13} 
\multicolumn{1}{c}{} &
  PIFu* \cite{saitoPIFuPixelAlignedImplicit2019}&
  \ding{53} &
  \ding{53} &
  2.525 &
  1.905 &
  0.155 &
  4.143 &
  2.773 &
  0.202 &
  3.603 &
  2.484 &
  0.186 \\
\multicolumn{1}{c}{} &
  PaMIR* \cite{zhengPaMIRParametricModelConditioned2020}&
  \ding{53} &
  \ding{53} &
  1.517 &
  1.331 &
  0.098 &
  1.768 &
  1.450 &
  0.102 &
  1.684 &
  1.410 &
  0.101 \\
\multicolumn{1}{c}{} &
  ICON \cite{xiuICONImplicitClothed2022}&
  \ding{53} &
  \ding{53} &
  \btwo 0.775 &
  \btwo 0.715 &
  0.054 &
  1.004 &
  0.930 &
  0.063 &
  0.928 &
  \btwo 0.859 &
  0.060 \\
\multicolumn{1}{c}{} &
  ECON \cite{xiuECONExplicitClothed2022}&
  \ding{53} &
  \ding{53} &
  0.912 &
  0.907 &
  \bone 0.037 &
  \btwo 0.926 &
  \btwo 0.917 &
  \bone 0.037 &
   \btwo 0.921 &
   0.914 &
   \bone 0.037 \\ \shline
\multirow{3}{*}{B} &
  \multicolumn{1}{c|}{$\text{Ours}_\text{D-IF (w/o Rectifier)}$} &
  \ding{53} & 
  \ding{53} &
  0.976 &
  0.900 &
  0.064 &
  1.245 &
  1.124 &
  0.075 &
  1.155 &
  1.049 &
  0.071 \\
 &
  \multicolumn{1}{c|}{$\text{Ours}_\text{D-IF}$} &
  \ding{53} &
  \ding{53} &
  0.721 &
  0.691 &
  0.050 &
  0.921 &
  0.880 &
  0.057 &
  0.854 &
  0.817 &
  0.055 \\
 &
  \multicolumn{1}{c|}{$\text{Ours}_\text{D-IF}$} &
  $\checkmark$ & 
  \ding{53} &
  0.712 &
  0.698 &
  0.051 &
  0.900 &
  0.870 &
  0.060 &
  0.838 &
  0.813 &
  0.057 \\ \hline
\multirow{2}{*}{C} &
  \multicolumn{1}{c|}{$\text{Ours}_\text{L2}$} &
  $\checkmark$ &
  L2 for $\mu$ &
  0.833 &
  0.777 &
  0.065 &
  1.019 &
  0.947 &
  0.070 &
  0.957 &
  0.890 &
  0.068 \\
 &
  \multicolumn{1}{c|}{\begin{tabular}[c]{@{}l@{}}$\text{Ours}_\text{KL}$\end{tabular}} &
  $\checkmark$ &
  Constant $\sigma$ &
  0.708 &
  0.689 &
  0.051 &
  0.885 &
  0.857 &
  0.057 &
  0.826 &
  0.801 &
  0.055 \\ 
\end{tabular}
}
\vspace{0.5 em}
\caption{\textbf{Quantitative evaluation.} (A) performance w.r.t. SOTA; and the ablation studies of (B) Implicit distribution fields (D-IF), Smooth Occupancy, and (C) Uncertainty distribution loss.  Notably, the best two results are colored as \colorbox{bestcolor}{first}~\colorbox{secondbestcolor}{second}.
}
\vspace{0.5 em}
\label{table:1}
\end{table*}

\begin{table*}[ht]
\centering
\resizebox{1.0\linewidth}{!}{
\begin{tabular}{c|ccc|ccc|ccc}
\multicolumn{1}{c|}{} & \multicolumn{3}{c|}{CAPE-FP} & \multicolumn{3}{c|}{CAPE-NFP} & \multicolumn{3}{c}{CAPE}   \\ 
Method &
  \multicolumn{1}{c}{Chamfer $\downarrow$} &
  \multicolumn{1}{c}{P2S $\downarrow$} &
  \multicolumn{1}{c|}{Normals $\downarrow$} &
  \multicolumn{1}{c}{Chamfer $\downarrow$} &
  \multicolumn{1}{c}{P2S $\downarrow$} &
  \multicolumn{1}{c|}{Normals $\downarrow$} &
  \multicolumn{1}{c}{Chamfer $\downarrow$} &
  \multicolumn{1}{c}{P2S $\downarrow$} &
  \multicolumn{1}{c}{Normals $\downarrow$} \\ 
  \shline
PIFu* \cite{saitoPIFuPixelAlignedImplicit2019}                 & 2.525  & 1.905 & 0.155 & 4.143   & 2.773 & 0.202  & 3.603 & 2.484 & 0.186 \\
$\text{PIFu}_\text{dis}$               & 1.867 (\textcolor{Green}{\textbf{-26\%}})   & 1.263 (\textcolor{Green}{\textbf{-34\%}})  & 0.111 (\textcolor{Green}{\textbf{-28\%}}) & 3.413 (\textcolor{Green}{\textbf{-18\%}})   & 1.985 (\textcolor{Green}{\textbf{-28\%}}) & 0.169 (\textcolor{Green}{\textbf{-16\%}})  & 2.898 (\textcolor{Green}{\textbf{-27\%}}) & 1.744 (\textcolor{Green}{\textbf{-30\%}}) & 0.150 (\textcolor{Green}{\textbf{-19\%}}) \\ \hline
PaMIR* \cite{zhengPaMIRParametricModelConditioned2020}                & 1.517 & 1.331 & 0.098 & 1.768   & 1.450 & 0.102  & 1.684 & 1.410 & 0.101 \\
$\text{PaMIR}_\text{dis}$               & 1.421 (\textcolor{Green}{-6\%})  & 1.175 (\textcolor{Green}{-12\%}) & 0.093 (\textcolor{Green}{-5\%}) & 1.612 (\textcolor{Green}{-9\%})  & 1.300 (\textcolor{Green}{-10\%})  & 0.092 (\textcolor{Green}{-10\%}) & 1.548 (\textcolor{Green}{-8\%}) & 1.258 (\textcolor{Green}{-11\%}) & 0.092 (\textcolor{Green}{-9\%}) \\ \hline
ICON \cite{xiuICONImplicitClothed2022}                  & 0.775   & 0.715 & 0.054 & 1.004   & 0.930 & 0.063  & 0.928 & 0.859 & 0.060 \\
$\text{ICON}_\text{dis}$                & 0.723 (\textcolor{Green}{-7\%}) & 0.713 (\textcolor{Green}{-0.2\%}) & 0.052 (\textcolor{Green}{-4\%}) & 0.900 (\textcolor{Green}{-10\%})  & 0.877 (\textcolor{Green}{-6\%}) & 0.060 (\textcolor{Green}{-5\%})  & 0.841 (\textcolor{Green}{-9\%}) & 0.822 (\textcolor{Green}{-4\%}) & 0.058 (\textcolor{Green}{-3\%}) \\ 
\end{tabular}}
\vspace{0.5 em}
\caption{\textbf{Generalizability proof on CAPE.} As a ``plug-and-play'' module, we apply implicit distribution fields (w/o smooth occupancy or uncertainty loss) to other methods, denoted as ``$X_\text{dis}$''. We denote the ``simulated'' SOTA methods with *.}
\label{table:2}
\end{table*}

\subsection{Representing mesh with smooth occupancy}
\label{sec: representing mesh with smooth occupancy}

\yihao{Following the above discussions, we tend to supervise the distribution of implicit values, like the occupancy field, by considering the relation between the relative location of query points and the associated uncertainty. To achieve this objective, an appropriate data structure informing the relative location of query points is imperatively. Conventional occupancy is typically binary, depicting whether a voxel is occupied by the interior of a surface or not. The binarity forgets all the quantitative information of the relative locations, which limits the spatial cost of computations but sacrifices the accuracy, especially for complicated shapes with fine details. While taking an accurate SDF as the representation would like to be computationally expensive when dealing with large-scale or high-resolution datasets. Moreover, neither of them is differentiated at points on the ground-truth surface. Therefore, in this paper, we adopt the smooth occupancy field to express a surface implicitly, which is also defined as the density field ~\cite{or2022stylesdf} in some contexts.}

\yihao{In particular, for a closed orientable surface $M\subset\mathbb{R}^3$ and a query point $p \in \mathbb{R}^3$, the spatial-aware smooth occupancy value $\mathit{O}(p|M)$ of $M$ at $p$ is defined as:
\begin{equation}\label{def：Smooth Occupancy}
\mathit{O}_{gt}(p)=\mathit{O}_{\alpha}(p|M):= \frac{1}{1+\exp(-\alpha\cdot{\rm SDF}(p|M))},
\end{equation}
where ${\rm SDF}(p|M)$ is the signed distance from $p$ to $M$ and $\alpha$ is the hyper-coefficient controlling the graininess of the voxel, where 
$\mathit{O}_{\infty}$ degenerates into the classical discrete (binary) occupancy field when $\alpha\to\infty$. 
It is worth mentioning the coefficient $\alpha$ can be either determined by priors or settled as the learnable parameter for training.
}

\yihao{Smooth occupancy field provides a smooth mapping $\mathit{O}_{\alpha}(\cdot|M):\mathbb{R}^3\mapsto [0,1]$ with the level set $\mathit{O}_{\alpha}^{-1}(0.5) = M$, which endows the implicit value of points near the surface $M$ with smoother gradients. In contrast to vanilla occupancy fields and signed distance functions (SDF), the smooth occupancy field exhibits differentiability and provides enhanced distance information. Finally, we apply Marching Cube~\cite{lorensen1987marching} on the learned smooth occupancy field at the level set of $\mathit{O}_{\alpha}^{-1}(0.5)$, to extract the triangle mesh.}

\subsection{Uncertainty distribution loss}
\label{sec: uncertainty distribution loss}

\begin{figure*}[t]
\vspace{-2.0 em}
   \includegraphics[trim=000mm 000mm 000mm 010mm, clip=True, width=1.0\linewidth]{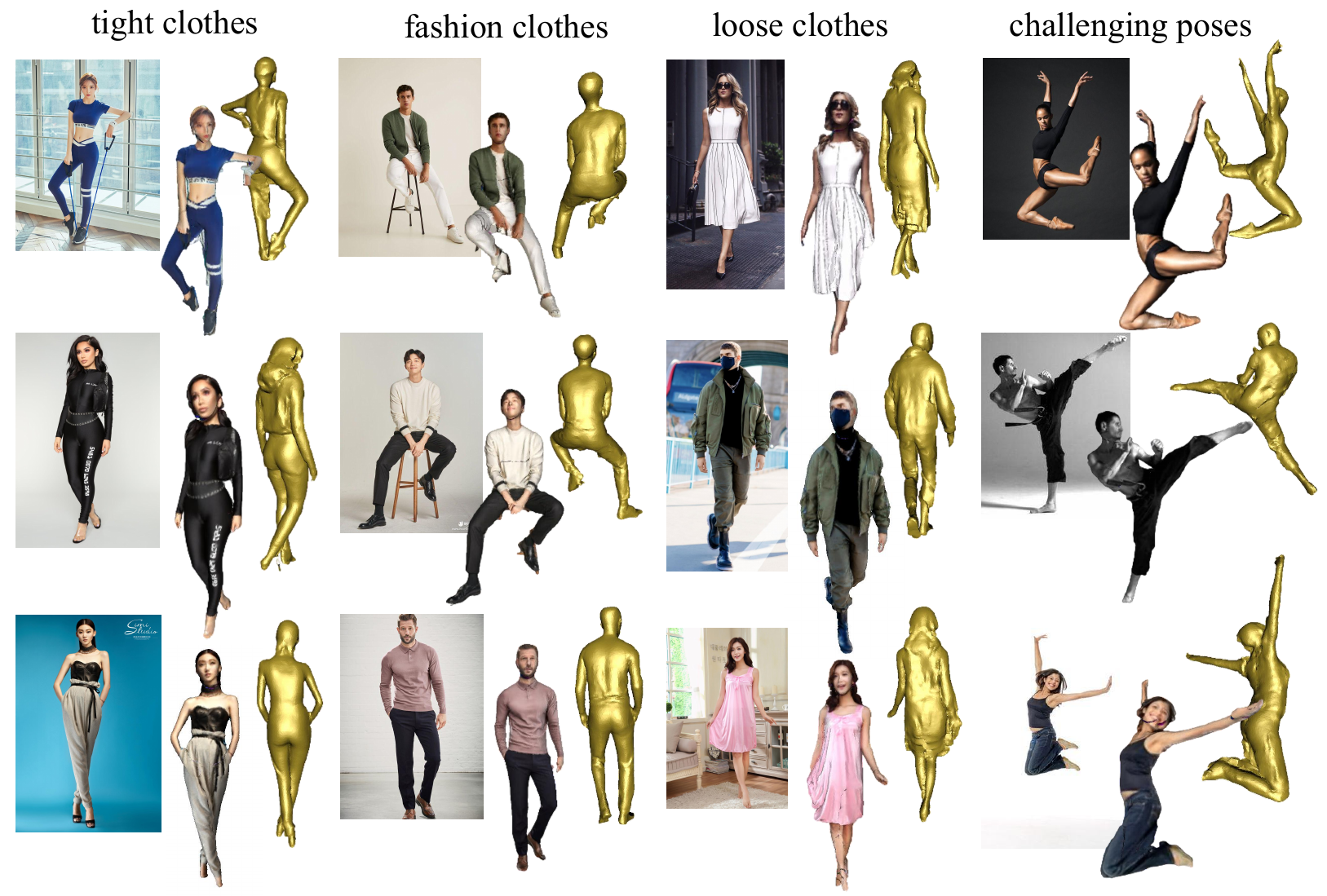}
   \caption{D-IF results for in-the-wild images with various clothing and challenging poses. \faSearch~\textbf{Zoom in} to see the geometric details.}
\label{fig:wild}
\end{figure*}


Training the D-IF network is nontrivial due to the necessity of effectively learning about uncertainty. With the smooth occupancy field, we can establish a connection between the variance of the distribution and spatial information through the uncertainty distribution loss.

Merely employing L1 loss to supervise the predicted occupancy distribution $P_{\varphi}\left(p\right)$ with the ground-truth smooth occupancy $\mathit{O}_{gt}(p)$ can yield accurate reconstructions by enforcing a sharp peak around the mean $\mu_{\varphi}\left(p\right)$. However, neglecting the supervision of variances may cause the network to converge to the Dirac impulse function~\cite{dirac1928quantum}. Therefore, we incorporate uncertainty into the learning process in our design, which encourages the network to learn randomness. Meanwhile, we attempt to involve a relatively higher uncertainty in the coarse prediction, which exploits the capacity of the Occupancy Rectifier to regress a finer result.



To address this, we design an uncertainty distribution loss $\mathcal{L}_\text{dis}$ to balance appropriate uncertainty and more accurate sampling values where we carefully consider their interplay. Specifically, we design an expected distribution $\hat{P}\left(p\right) =\mathcal{N}\left({\mu_d}{(p)}, \sigma_{d}{(p)}\right)$ with the a mean and variance $\mu_d$ and $\sigma_{d}$ by taking the point-to-mesh distance into account. 
The mean of the designed distribution ${\mu_d}{(p)}$ for the query point is defined naturally as the ground-truth smooth occupancy $\mathit{O}_{gt}(p)$.
As illustrated in~\cref{fig:pipeline}, $\sigma_{d}{(p)}$ is designed to be positively correlated with the point-to-mesh distance by
\begin{equation}
\yihao{\sigma_{d}{(p)} = k e^{-\beta{\left(\mu-0.5\right)}^2},}
\label{eq:gt-sigma}
\end{equation}
where $k$ and $\beta$ are hyperparameters to control the behavior of the uncertainty. 

Consequently, we adopt the KL-divergence~\cite{liese2006divergences} to measure the distribution disparity from the perdition to the designed ground truth,
\begin{equation}
\mathcal{L}_\text{dis} = {\rm KL}(P_{\varphi}\left(p\right) \mid \mid \hat{P}\left(p\right)).
\label{eq:kl-loss}  
\end{equation}
The total uncertainty loss of our network $\mathcal{L}_\text{un}$ is the combination of the distribution and 3D reconstruction losses:
\begin{equation}
\begin{aligned}
&\mathcal{L}_\text{rec} = {\mid \mid \widetilde{O_s}(p) - O_{gt}(p) \mid \mid} ^ 2,\\
&\mathcal{L}_\text{un} = \alpha_1\mathcal{L}_\text{dis} + \alpha_2\mathcal{L}_\text{rec}.
\end{aligned}
\end{equation}
where $\alpha_1$, $\alpha_2$ are fixed loss weighting factors.


\section{Experiment}

\begin{figure*}[h]
\begin{center}
   \includegraphics[width=1.0\linewidth]{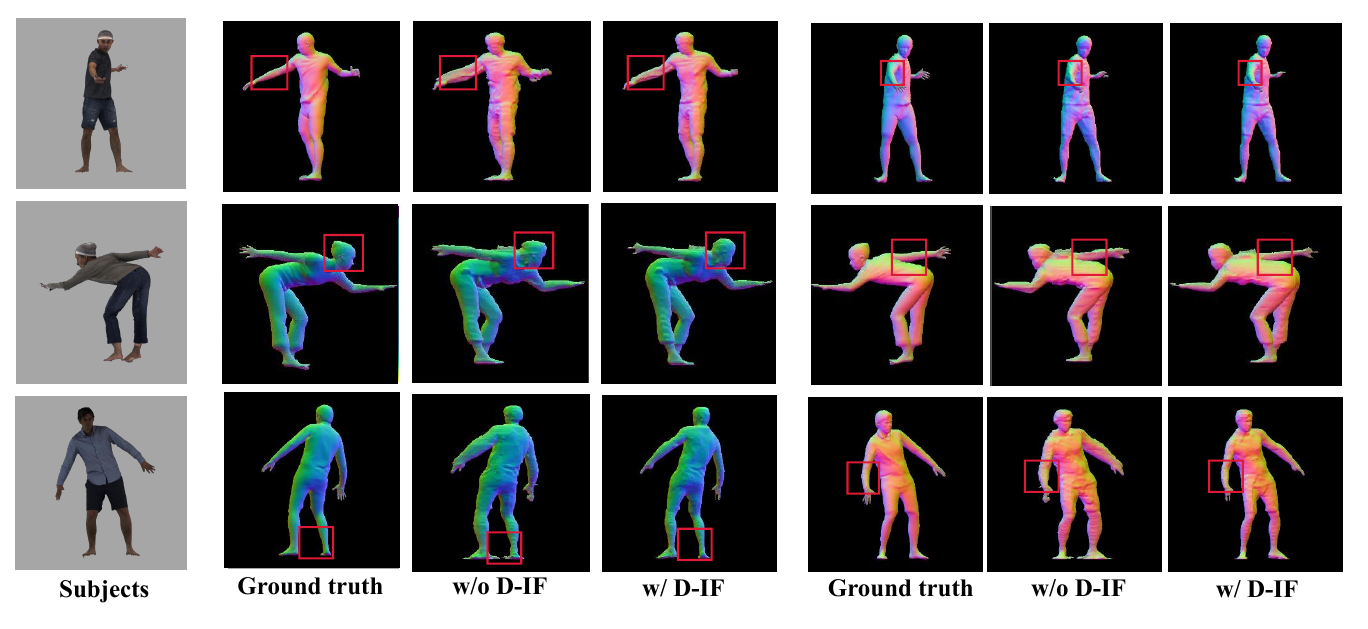}
\end{center}
 \vspace{-0.2in}
   \caption{Reconstructions w/ and w/o implicit distribution field (D-IF). \faSearch~\textbf{Zoom in} to see the geometric details.}
\label{fig:7}
\end{figure*}

\qheading{Training} All the baselines are trained on THuman2.0~\cite{yu2021function4d}, which contains 525 high-quality human-textured scans in various poses captured by a dense DSLR rig along with their corresponding SMPL-X fits. The model is firstly trained with $\mathcal{L}_\text{rec}$ for 10 epochs, the learning rate is $1 \times 10^{-4}$. Then, we replace $\mathcal{L}_\text{rec}$ with uncertainty loss $\mathcal{L}_\text{un}$, and fine-tune the model for 5 epochs with the same learning rate. Empirically, we set $\alpha_1 = 1.0$, $\alpha_2 = 0.55$, $k = 0.6$ and $\beta = 4$. A larger $q$ leads to a higher level of accuracy in the reconstruction process. However, an extremely large $q$ will degenerate the smooth occupancy into a discrete 0-1 occupancy, which can harm the quality due to the loss of distance information. So we take a balance and set $q=1 \times 10^{3}$. The entire training process takes approximately 1 day on a single NVIDIA RTX 3090 GPU with 36.5M learning parameters.

\qheading{Testing} We mainly follow the test setting of ICON~\cite{xiuICONImplicitClothed2022}, which totally selects 150 scans from CAPE~\cite{ma2020learning} to evaluate the reconstruction accuracy under challenging poses (``CAPE-NFP'', 100 scans), and fashion poses (``CAPE-FP'', 50 scans). The testing RGB images are rendered by rotating a virtual camera around the textured scans by \{$0^{\circ}$, $120^{\circ}$, $240^{\circ}$ \}. During the evaluation, we randomly sample the coarse occupancy value from the predicted distribution into Occupancy Rectifier to reconstruct the final prediction.

\paragraph{Metrics.} ``Chamfer'' and ``P2S'' mainly capture the coarse geometric errors, while ``Normals'' mainly measure the high-frequency difference:

\begin{itemize}
\itemsep 0.1em
\item {\bf P2S distance (cm)}. The point-to-surface (P2S) distance is computed from randomly sampled scan points to their closest face on the reconstructed mesh. 
\item {\bf Chamfer distance (cm)}. This metric could be regarded as \textit{bilateral} P2S distance, which also considers the P2S distance from randomly sampled points on the reconstructed mesh to the ground-truth scans.
\item {\bf Normals difference (L2)}. The normal images used for evaluation are rendered from the reconstructed and ground-truth meshes, by rotating a virtual camera around them by \{ $0^{\circ}$, $90^{\circ}$, $180^{\circ}$, $270^{\circ}$ \}. 
\end{itemize}



\subsection{Comparison with state-of-the-art methods.} 

\begin{figure*}[t]
\vspace{-3.0 em}
   \includegraphics[width=1.0\linewidth]{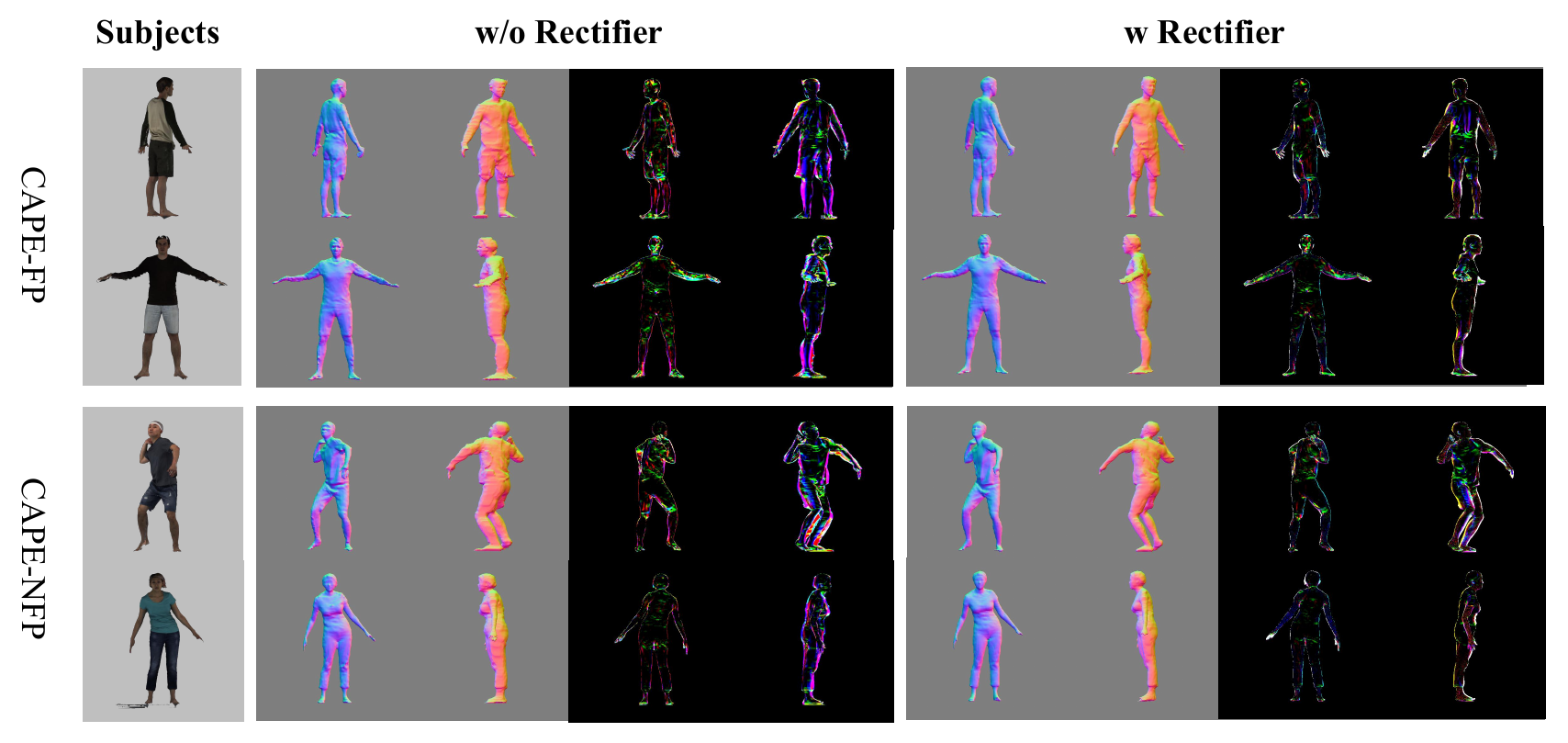}
\vspace{-1.5 em}
   \caption{Error map for the reconstructed mesh w/ and w/o Occupancy Rectifier. \faSearch~\textbf{Zoom in} to see the geometric details.}
\label{fig:8}
\end{figure*}

We conduct quantitative comparisons with mainstream SOTAs, including PIFu~\cite{saitoPIFuPixelAlignedImplicit2019}, PaMIR~\cite{zhengPaMIRParametricModelConditioned2020}, ICON~\cite{xiuICONImplicitClothed2022} and ECON~\cite{xiuECONExplicitClothed2022}. For a fair comparison, we follow the re-implementations of PIFu and PaMIR in ICON~\cite{xiuICONImplicitClothed2022}, and retrain all models on THuman2.0 ~\cite{yu2021function4d}.


The quantitative comparison results are presented in~\cref{table:1}-A. Our proposed method has demonstrated superior performance compared to all baselines, confirming its superiority by a significant margin. Notably, our method has achieved even better results than the previous best-performing method, ECON~\cite{xiuECONExplicitClothed2022}, on the challenging ``CAPE-NFP'' dataset. This dataset evaluates the model's ability to reconstruct clothed humans with out-of-distribution poses. The impressive performance of our method on this challenging dataset highlights its robustness and effectiveness in handling complex real-world scenarios.


The results in~\cref{fig:0} clearly demonstrate that our approach produces more accurate and detailed reconstructions. This aligns with our objective of using uncertainty distribution instead of deterministic occupancy values to represent the geometries. The superiority of our method can be attributed to the innovative point-wise distribution learning, supplemented by the inclusion of uncertainty loss.

In summary, these two findings discussed above provide strong evidence that uncertainties exist in the clothed human reconstruction process and that they follow explainable laws. These observations confirm the validity of our motivation to leverage point-wise distribution learning to better capture uncertainty. By providing a more accurate representation of uncertainty in the reconstruction process, our approach has the potential to improve the quality and reliability of clothed human mesh reconstruction. Overall, our results support the use of uncertainty learning in this context and highlight the importance of considering uncertainty in machine learning applications. More reconstruction results for in-the-wild images are shown in~\cref{fig:wild}.

\subsection{Ablation Study.} We further validate the effectiveness of 1) implicit distribution field (D-IF) in~\cref{table:1}-B, 2) Occupancy Rectifier in~\cref{table:3}, and 3) uncertainty distribution loss in~\cref{table:1}-C.

\medskip
\qheading{Implicit Distribution Field (D-IF)} As shown in~\cref{table:1}-B, we replace ``Value-based Implicit Function (IF)'' used by competitors with our proposed ``Implicit Distribution Field (D-IF)'', and benchmark their performance under original occupancy / smooth occupancy. We find that uncertainty distribution has significantly improved the reconstruction quality, especially for stretching gestures, and also reduce the abnormal artifacts of limbs; see~\cref{fig:7}. 

\begin{table}[!h]
\centering
\resizebox{\linewidth}{!}{
\begin{tabular}{l|c|cc|cc}
   &
  \begin{tabular}[c]{@{}c@{}} Rectifier\end{tabular} &
  P2S (cloth) $\downarrow$ &
  Chamfer (cloth) $\downarrow$ &
  P2S (body) $\downarrow$ &
  Chamfer (body) $\downarrow$ \\ \shline
\multirow{2}{*}{Loose}&
  \ding{53} &
  1.033 &
  1.090 &
  1.268 &
  1.377 \\
   &
  \checkmark & 
  0.955 (\textcolor{Green}{-7\%}) &
  0.927 (\textcolor{Green}{-14\%}) &
  0.843 (\textcolor{Green}{-33\%}) &
  0.864 (\textcolor{Green}{-37\%}) \\ \hline
\multirow{2}{*}{Tight}&
  \ding{53} &
  0.910 &
  0.988 &
  0.939 &
  1.035 \\
 &
  \checkmark &
  0.586 (\textcolor{Green}{-35\%}) &
  0.607 (\textcolor{Green}{-38\%}) &
  0.639 (\textcolor{Green}{-31\%}) &
  0.669 (\textcolor{Green}{-35\%}) \\
\end{tabular}
}
\vspace{0.5 em}
\caption{\textbf{Ablation experiments (w/ Rectifier vs. w/o Rectifier) on both loose and tight clothing.} Metrics (cloth) are calculated between the reconstructed surface and ground truth \textit{clothed} surface, and metrics (body) are calculated between the reconstructed surface and ground truth \textit{body} surface. The results indicate that the Occupancy Rectifier enhances reconstruction accuracy mainly by bringing the predicted mesh closer to the SMPL body mesh.}
\label{table:3}
\end{table}

\smallskip
\qheading{Occupancy Rectifier} Besides, we also demonstrate the necessity of Occupancy Rectifier in~\cref{table:1}-B. Results show that removing the Rectifier ($\text{Ours}_\text{D-IF (w/o Rectifier)}$) finally harms the reconstruction quality, increasing Chamfer error from 0.854 cm to 1.155 cm ($\textcolor{red}{35 \%}$ increase), and see normal error maps (w/ Rectifier vs. w/o Rectifier) in~\cref{fig:8}. 

To further validate the impact of the Occupancy Rectifier, we conducted ablation studies (w/ vs w/o Rectifier) on both loose and tight clothing categories in the CAPE dataset. The categorization was based on the ``P2S'' and ``chamfer'' distances (measured in centimeters) between the reconstructed surface and ground truth clothed mesh or minimally-clothed mesh. The evaluation results presented in~\cref{table:3} demonstrate that, for subjects wearing loose clothing, there were significant reductions in P2S (body, \textcolor{Green}{33\%} decrease) and Chamfer (body, \textcolor{Green}{37\%} decrease) when employing the Occupancy Rectifier. However, there were comparatively smaller decreases observed in P2S (cloth \textcolor{Green}{7\%} decrease) and Chamfer (cloth, \textcolor{Green}{14\%} decrease). For subjects wearing tight clothing, both Chamfer (body, \textcolor{Green}{35\%} decrease) and Chamfer (cloth, \textcolor{Green}{38\%} decrease) exhibited substantial improvements. These findings suggest that the Rectifier primarily enhances reconstruction accuracy by bringing the predicted mesh closer to the SMPL body mesh.

\smallskip
\qheading{Uncertainty Distribution Loss} Finally, as shown in~\cref{table:1}-C, we compare the uncertainty distribution loss to other versions, which simply regress the predicted $\mu$ with L2 loss or change the variance of the target distribution into a constant (flat distribution). The chosen constant is the surface value of the designed distribution, in~\cref{eq:gt-sigma}. The results show that the introduction of the designed distribution in uncertainty loss does improve the robustness of reconstruction, as described in~\cref{sec: uncertainty distribution loss}.

\subsection{Generalization of D-IF.} The core module, Implicit Distribution Field (D-IF), can be easily integrated into other implicit-based approaches, like PIFu~\cite{saitoPIFuPixelAlignedImplicit2019}, PaMIR~\cite{zhengPaMIRParametricModelConditioned2020} and ICON~\cite{xiuICONImplicitClothed2022}. We re-train the above three baselines with our D-IF, see~\cref{table:2} for the quantitative comparison. D-IF boosts all the baselines on all metrics, especially for PIFu, where D-IF even reduces the reconstruction error by \textcolor{Green}{27\%} in Chamfer, \textcolor{Green}{30\%} in P2S, and \textcolor{Green}{19\%} in Normals. Additionally, by incorporating uncertainty, D-IF improves the recovery of non-rigid deformations in clothing, as shown in non-fashion poses (NFP) in~\cref{table:2}. All these results have demonstrated that D-IF generalizes quite well to a wide range of implicit-based human reconstruction approaches. Hence, we have reason to believe that D-IF, as a powerful \textit{plug-and-play} module, could also be used to represent other non-human shapes.

\section{Conclusion}
Clothed human reconstruction is a fundamental task in human digitization. In this paper, we propose a novel method to learn the uncertainty that exists in clothed human reconstruction process for better geometric details. In our work, a new implicit representation named smooth occupancy field is introduced to represent clothed humans in neural space. Then, a distribution-guided implicit function network is proposed to learn point-wise distribution to recover the occupancy distribution field. Finally, a novel uncertainty loss is presented to better train the network. Experimental results demonstrate that our method achieves state-of-the-art performance and can be generalized to other baselines to consistently improve their performances. 


\bigskip
\qheading{Acknowledgments} 
Xueting Yang, Zhaoxin Fan, Wei Wang and Hao Xu are supported by \href{https://www.psyai.com/home}{Psyche AI Inc}. Yihao Luo is partly funded by the Imperial College London and partly supported by  Psyche AI Inc. Yuliang Xiu is funded by the European Union’s Horizon $2020$ research and innovation programme under the Marie Skłodowska-Curie grant agreement No.$860768$ (\href{https://www.clipe-itn.eu}{CLIPE}).


\clearpage
{\small
\bibliographystyle{ieee_fullname}
\bibliography{egbib}
}

\end{document}